\documentclass[journal]{IEEEtran}

\usepackage[pdftex]{graphicx}
\usepackage{amssymb} 
\usepackage{graphicx}
\usepackage[natural]{xcolor}
\usepackage[markup=underlined]{changes}
\usepackage{todonotes}
\usepackage[cmex10]{amsmath}
\usepackage{mathtools}
\usepackage[tight,footnotesize]{subfigure}
\usepackage{url}
\usepackage{algorithm}
\usepackage{algorithmicx}
\usepackage[noend]{algpseudocode}
\usepackage{gensymb}
\usepackage{siunitx} 
\usepackage{interval}

\definechangesauthor[name=Stefan, color=blue]{sz}

\begin{document}

\title{Cyclist Intention Detection:\\
	A Probabilistic Approach}


\author{Stefan Zernetsch, Hannes Reichert, Viktor Kress, Konrad Doll,~\IEEEmembership{Member,~IEEE,} and Bernhard Sick~\IEEEmembership{Member,~IEEE,}
	\thanks{S. Zernetsch, H. Reichert, V. Kress, and K. Doll are with the Faculty of Engineering,
		University of Applied Sciences Aschaffenburg, Aschaffenburg, Germany
		{\tt\footnotesize stefan.zernetsch@th-ab.de,
			hannes.reichert@th-ab.de, viktor.kress@th-ab.de konrad.doll@th-ab.de}}
	\thanks{B. Sick is with the Intelligent Embedded Systems Lab, University of Kassel,
		Kassel, Germany
		{\tt\footnotesize bsick@uni-kassel.de}}
}

\maketitle

\begin{abstract}
	
This article presents a holistic approach for probabilistic cyclist intention detection. A basic movement detection based on motion history images (MHI) and a residual convolutional neural network (ResNet) are used to estimate probabilities for the current cyclist motion state. These probabilities are used as weights in a probabilistic ensemble trajectory forecast. The ensemble consists of specialized models, which produce individual forecasts in the form of Gaussian distributions under the assumption of a certain motion state of the cyclist (e.g. cyclist is starting or turning left). By weighting the specialized models, we create forecasts in the from of Gaussian mixtures that define regions within which the cyclists will reside with a certain probability. To evaluate our method, we rate the reliability, sharpness, and positional accuracy of our forecasted distributions. We compare our method to a single model approach which produces forecasts in the form of Gaussian distributions and show that our method is able to produce more reliable and sharper outputs while retaining comparable positional accuracy. Both methods are evaluated using a dataset created at a public traffic intersection. Our code and the dataset are made publicly available.

\end{abstract}

%
\IEEEpeerreviewmaketitle



\section{\large Introduction}
\label{sec_introduction}
\subsection{Motivation}

In future traffic scenarios, more and more automated vehicles will share the road with vulnerable road users (VRU), such as pedestrians and cyclists. To guarantee safe interaction, automated vehicles not only have to be able to capture the current traffic environment, but they must also anticipate future movements of all traffic participants. While other automated vehicles can share their planned paths, the future trajectories of VRU have to be estimated. However, since VRU can change their path quickly, trajectory forecasts are error prone. Therefore, instead of deterministic positional forecasts, we have to estimate regions within which the VRU will reside with a certain probability. In this article, we present a method to forecast VRU positions in the form of probability distributions using a weighted ensemble forecast. By weighting outputs in the form of Gaussian distributions created by motion type specific Bayesian neural networks, we are able to reduce forecast errors and forecast more reliable and sharper distributions compared to a single network. The used weights are created by a basic movement detection method, based on motion history images (MHI) \cite{ahad_mhi}, which encompass the cyclist movements over a short period of time within a single image. The method distinguishes between certain motion states, such as \textit{wait} or \textit{start}. Since we want to give statistical guarantees that the VRU resides within the forecasted regions, we have to evaluate the reliability of the network outputs. Building on a previous approach for forecasted normal distributions, we present a universal method to evaluate the reliability of forecasted probability distributions. In addition to reliability, sharpness, i.e., narrowness, of the forecasts is crucial for successful path planing since a very wide, but reliable forecast could prohibit a planning algorithm from choosing a safe path around the cyclist. Hence, we present a method to evaluate the sharpness of forecasts.

\subsection{Related Work}

The field of VRU intention detection has become an active field of research over the past years. The developed methods can roughly be grouped into basic movement detection, where VRU motion states such as \textit{wait} or \textit{start} ought to be detected, and trajectory forecast, with the goal to anticipate future VRU positions.
Publications regarding basic movement detection often focus on a specific scenario or movement type. In \cite{willpedcross}, four methods based on Gaussian process dynamical models (GPDM) and features derived from dense optical flow to detect pedestrians' intention to stop at a curbside or continue walking were proposed. GPDM in combination with human poses extracted from image sequences to detect start intentions of pedestrians were introduced in \cite{Quintero.2017} and \cite{Quintero.2019}. In \cite{kress.itsc.2019}, the authors used 3D poses of cyclists to detect starting motions from a moving vehicle. In \cite{Koehler.2013} and \cite{koehler2015}, an approach based on MHI in combination with a support vector machine (SVM) to detect pedestrians' starting and stopping intentions was presented. In both cases, the MHI based approach outperformed an approach based on a Kalman filter with interacting multiple models. In \cite{ownZernetsch_iv2018}, the MHI based approach was adapted for cyclist starting movement detection and extended by a residual convolutional neural network (ResNet), where the ResNet outperformed the SVM approach. In this article, the approach from \cite{ownZernetsch_iv2018} is adapted to a wide angle stereo camera system and extended to cover all possible movement types of cyclists.

VRU trajectory forecast can be grouped into deterministic and probabilistic forecast methods. Deterministic approaches usually rely on past VRU positions to forecast future positions without quantization of uncertainties. For example, in \cite{quintero2018}, the authors used GPDM with human poses to forecast pedestrian trajectories up to one second into the future. In \cite{Goldhammer.2015}, the authors proposed a method to forecast pedestrian trajectories using polynomial approximation of the pedestrians' past velocity in combination with a multi layer perceptron (PolyMLP). The approach is extended to cyclist trajectories and compared to two physical models in \cite{ownZernetsch_iv2016}, where the MLP outperformed both approaches. In \cite{ownBieshaarTrans2018} and \cite{goldhammer2019}, basic movement detections are used to improve the accuracy of trajectory forecasts by training specialized forecast models and weighting the outputs with pseudo-probabilistic outputs from basic movement detections. By using the PolyMLP approach from \cite{Goldhammer.2015} in combination with the MHI based approach from \cite{koehler2015}, the authors were able to slightly improve the positional accuracy of pedestrian trajectory forecasts. In \cite{bieshaar2017itsc} a cooperative basic movement detection approach based on smart device sensor data and measurements from a camera network was used. By weighting two specialized forecasting models based on \cite{Goldhammer.2015}, the authors were able to reduce the forecast error of starting motions by 21\%.

Although these methods show promising results, to serve as basis for motion planning algorithms in automated vehicles, the uncertainty of positional forecast needs to be estimated. This is often achieved through prediction of regions instead of positions in  the form of probability distributions. In \cite{Eilbrecht.2017}, the method from \cite{Goldhammer.2015} was enhanced by an uncertainty estimate created by an unconditional model. The uncertainty estimates were used with a planning method based on predictive control. In \cite{Koschi2018a}, the authors performed set-based predictions using reachability analysis that provide bounded regions that include all possible future positions of pedestrians ensuring the safety
of planned motions. In \cite{Alahi_2016_CVPR}, the authors proposed a method to forecast trajectories of pedestrians in a crowded area, taking into account the interaction between different pedestrians. A long short-term memory model was used to forecast positions in the from of bivariate Gaussian distributions, the model was trained by minimizing the negative log-likelihood. Recurrent neural networks in combination with context information to forecast cyclist position in the form of Gaussian distributions were used in \cite{pool_iv2019}. While all the methods mentioned above create probabilistic forecasts, the reliability of the produced outputs is not verified. In \cite{ownZernetsch_iv2019}, a similar approach to forecast cyclist trajectories is used and a method to evaluate the reliability of the forecasted distributions is proposed. However, the method is limited to unimodal Gaussian distributions. The article also shows that modeling future trajectories with unimodal Gaussian distributions leads to underconfident estimations. In this work, we extend the approach from \cite{ownZernetsch_iv2019}. By performing basic movement detection to generate weights for different movement types and training movement specific forecast models, we are capable of forecasting multimodal distributions. Furthermore we extend the evaluation approach from \cite{ownZernetsch_iv2019} for arbitrary distributions and show that our approach outperforms \cite{ownZernetsch_iv2019} in terms of reliability.

In addition to reliability, which assesses the correctness of forecasted probability densities, we need to evaluate sharpness, i.e., narrowness, of the forecasted distributions. To the best of our knowledge, there has been no evaluation of sharpness of distributions in the field of intention detection so far. The concept of sharp probability distributions was described in \cite{sharpness_nielsen}. We adapted the method to our approach.

While there has been some research in the field of pedestrian intention detection, cyclists are mostly neglected in the field. Therefore, we focus on the detection of cyclist intention, however, our approaches can be transfered to other VRU as well.

\subsection{Main Contributions and Outline of this Paper}

Our main contribution is a holistic approach for probabilistic cyclist intention detection. By combining a basic movement detection with a probabilistic trajectory forecast, we are able to produce reliable and sharp estimates about the possible future positions of cyclists. Our basic movement detection method creates two-channel MHI using instance segmentations, capturing the movement of cyclist and bicycle separately. In addition, we model all possible movements of cyclists. The probabilistic trajectory forecast is based on a mixture of Gaussians by creating specialized forecasting models for different movement types and weighting their outputs using the probabilities of the basic movement detection. To evaluate our results, we developed two methods which are capable of rating the reliability and sharpness of arbitrary distributions. By using the method from \cite{ownZernetsch_iv2019} as baseline, we show that our algorithm is able to produce more reliable and sharper forecasts with comparable positional accuracy. Both methods are evaluated using a dataset, which we created at an urban intersection in real world traffic scenarios. The dataset and our code is made publicly available \cite{zernetsch_github_tiv}\cite{zernetsch_cyclistactionrec_mhi}.

The remainder of this article is structured as follows: In Sec. \ref{subsec:dataset}, the dataset and the research intersection where the dataset was created are described. Sec. \ref{subsec:intention_detection} outlines the overall intention detection process. Basic movement detection and probabilistic trajectory forecast are described in Sec.~\ref{subsec:basic_movement_detection} and Sec. \ref{subsec:probabilistic_trajectory_forecast}, respectively. In Sec. \ref{sec_ResultsOutline}, we evaluate the results of our trained models, followed by a conclusion and short outlook in Sec. \ref{sec:conclusion}.
\section{\large Method}
\label{sec_method_overview}

\subsection{Test Site and Dataset}
\label{subsec:dataset}

\begin{figure*}
	\begin{center}
		\includegraphics[width=\textwidth]{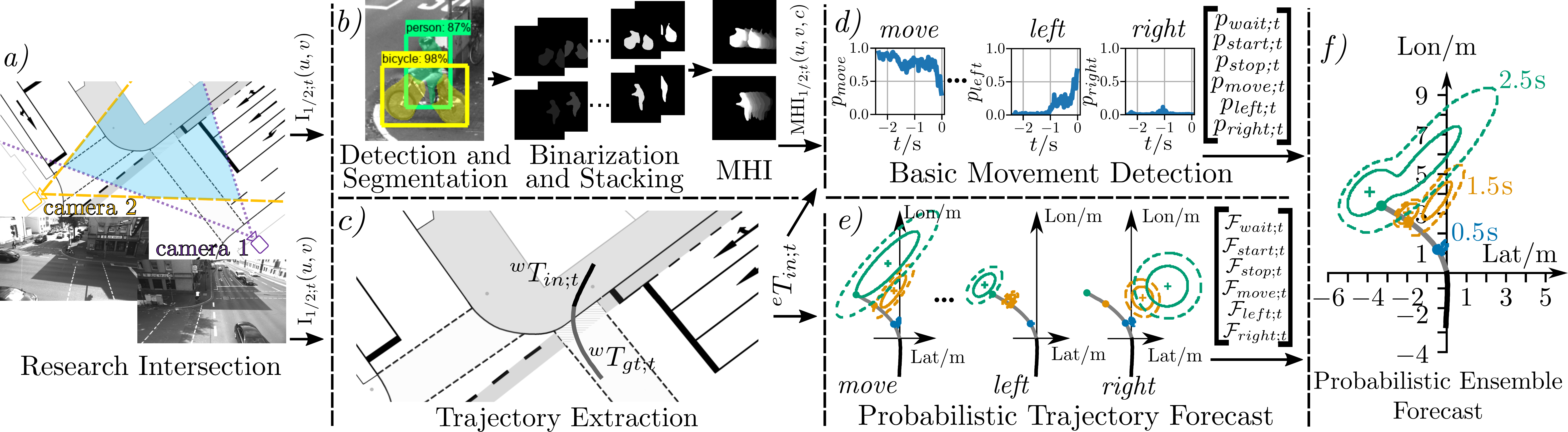}
		\vskip 0mm
		\caption{Probabilistic intention detection pipeline: a) Images $I_{1/2;t}(u, v)$ are captured using wide angle stereo camera sytem and passed to b) and c); b) Person and bicycle are segmented in images to create $MHI_{1/2;t}(u, v, c)$. MHI is passed to d); c) Cyclist head trajectories in ego coordinates $^eT_{in;t}$ are extracted and passed to d) and e). Ground truth trajectory $^eT_{gt;t}$ is passed to e) during training; d) Basic movement detection is performed. Resulting probabilities $p_{s;t}$ are passed as weights to f); e) Probabilistic trajectory forecast is performed individually for every specialized forecast model. Results in the form of density functions $\mathcal{F}_{s;t}$ are passed to f); f) Density functions $\mathcal{F}_{s;t}$ are weighted using $p_{s;t}$ to generate ensemble forecasts.}
		\label{fig:pipeline}
	\end{center}
	\vskip 0mm
\end{figure*}

In this section we provide a description of the test site where the dataset is created, as well as a description of the dataset itself.

The dataset was created at the research intersection of the University of Applied Sciences Aschaffenburg \cite{GoldhammerXung} using a wide angle stereo camera system (Fig. \ref{fig:pipeline}a). The intersection located at the university consists of four arms, two arms with 5 vehicle lanes each, one arm with three lanes, and one with two lanes. With up to 24,000 vehicles a day, the intersection is strongly frequented. To gather video data of VRU, a stereo camera system was mounted to capture a corner of the intersection with a side walk, two pedestrian crossings, and a bike lane. The full HD gray scale cameras 1 and 2 are mounted approx. \SI{5}{\meter} above ground at an angle of approx. \ang{90} relative to each other. Using this setup, most occlusions by vehicles or other VRU can be avoided. The cameras operate at 50 fps.

The dataset consist of scenes containing cyclists moving across the research intersection. We recorded 672 scenes where different cyclists were instructed to move between certain end points at the intersection, while following the traffic rules, and 976 scenes of uninstructed cyclists, resulting in 1,648 scenes in total. As basis for our developed methods, we recorded images $\text{I}_{1/2;t}(u,v)$ from cameras 1 and 2, with pixel positions $u$ and $v$, at the current time step $t$. These images are used to extract MHI and trajectories (described in Sec. \ref{subsec:intention_detection}), which we made publicly available \cite{zernetsch_cyclistactionrec_mhi}.
\subsection{Intention Detection}
\label{subsec:intention_detection}

In this section, we describe the overall intention detection process. We define intention detection as umbrella term which encompasses detection of motion states, such as \textit{wait} or \textit{start}, which we define as basic movement detection, and the forecast of the cyclist positions in the form of probability distributions, which we define as probabilistic trajectory forecast.

Fig. \ref{fig:pipeline} describes the intention detection process. In the first step (Fig. \ref{fig:pipeline}a), the images $\text{I}_{1/2;t}(u, v)$, with pixel positions $u$ and $v$, are grabbed from the cameras 1 and 2 and passed to the MHI generation (Fig.~\ref{fig:pipeline}b) and trajectory extraction (Fig.~\ref{fig:pipeline}c). 

To generate MHI, the cyclist is detected and tracked in both camera images. An instance segmentation is performed on both images to detect cyclists and bicycles. The intensity values of the cyclist and bicycle silhouettes created by instance segmentation from the past $M$ time steps are decayed over time and stacked resulting in $\text{MHI}_{1/2;t}(u, v, c)$, with two image channels $c$ containing separate motion histories of cyclists and bikes. 

For trajectory extraction, the head of the cyclist is detected in images of both cameras at every time step. The head detections are then triangulated in order to receive the head position $^w\vec{y}_{t}=[^wx_{t}, ^wy_{t}]$ in world coordinates (denoted by $^w$) $^wx_{t}$ and $^wy_{t}$ (Fig. \ref{fig:pipeline}c). As input for both basic movement detection and trajectory forecast, we use the past 50 head detections, resulting in a trajectory $^wT_{in;t}=\{^w\vec{y}_{t},^w\vec{y}_{t-\SI{0.02}{\second}},^w\vec{y}_{t-\SI{0.04}{\second}}, ...^w\vec{y}_{t-\SI{1}{\second}}\}$ consisting of the cyclist's head positions over an input time set $H_{\text{input}}=\{\SI{0}{\second},\SI{-0.02}{\second},\SI{-0.04}{\second},...,\SI{-1}{\second}\}$ over the past second. To train the trajectory forecast network, the future trajectory $^wT_{gt;t}=\{^w\vec{y}_{t+\SI{0.1}{\second}},^w\vec{y}_{t+\SI{0.2}{\second}}, ...^w\vec{y}_{t+\SI{2.5}{\second}}\}$ with a forecast time set $H_{\text{forecast}}=\{\SI{0.1}{\second},\SI{0.2}{\second},...,\SI{2.5}{\second}\}$, is used as ground truth. Before the trajectories are passed to the forecast model, a transformation of the positions from world to ego coordinates is performed:
\vskip -4mm
\begin{equation}
^e\vec{y}_{t+h} = \begin{bmatrix}
\cos(\varphi_t) & \sin(\varphi_t)\\
-\sin(\varphi_t) & \cos(\varphi_t)
\end{bmatrix}\cdot (^w\vec{y}_{t+h}-^w\vec{y}_{t}),
\label{equ:ego_transformation}
\end{equation} with $\varphi_t$ being the movement direction of the cyclist at the current time $t$ and $h\in H_{\text{input}}\cup H_{\text{forecast}}$ the relative time offset from t. The transformation is performed for both,  $^wT_{in;t}$ and $^wT_{gt;t}$, resulting in $^eT_{in;t}$ and $^eT_{gt;t}$, which are passed to the probabilistic trajectory forecast (Fig.~\ref{fig:pipeline}e). $^wT_{in;t}$ is passed to basic movement detection (Fig.~\ref{fig:pipeline}d).

Basic movement detection is performed using $\text{MHI}_{1;t}(u, v, c)$, $\text{MHI}_{2;t}(u, v, c)$, and $^eT_{in;t}$. It generates a probability estimate $p_{s;t}$ for every possible motion state $s$ at current time $t$ (Fig.~\ref{fig:pipeline}d).

The probabilistic trajectory forecasts of specialized models (Fig.~\ref{fig:pipeline}e) are used to generate positional forecasts in the form of Gaussian distributions $\mathcal{F}_{s;t}$. We use individual models, each specifically trained for one motion state $s$ of the cyclist.

In the final step (Fig.~\ref{fig:pipeline}f), the specialized forecasts $\mathcal{F}_{s;t}$ are weighted using basic movement detection probabilities $p_{s;t}$ and are combined to one single distribution.


\subsection{Basic Movement Detection}
\label{subsec:basic_movement_detection}

In this section, the detection of basic movements using MHI and past trajectories of cyclists is explained. In Sec.~\ref{subsec:basic_movement_detection_basic_movements}, we define the set of basic movements we aim to detect, Sec.~\ref{subsec:basic_movement_detection_mhigeneration} describes how the MHI are generated, in Sec.~\ref{subsec:basic_movement_detection_networkarch}, we present our detection method including the network architecture, and Sec.~\ref{subsec:basic_movement_detection_evaluation} describes the methods and metrics we used to evaluate the detection algorithms.

\subsubsection{Basic Movements}
\label{subsec:basic_movement_detection_basic_movements}

\begin{figure}
	\begin{center}
		\vskip 0mm
		\includegraphics[width = 0.7\columnwidth]{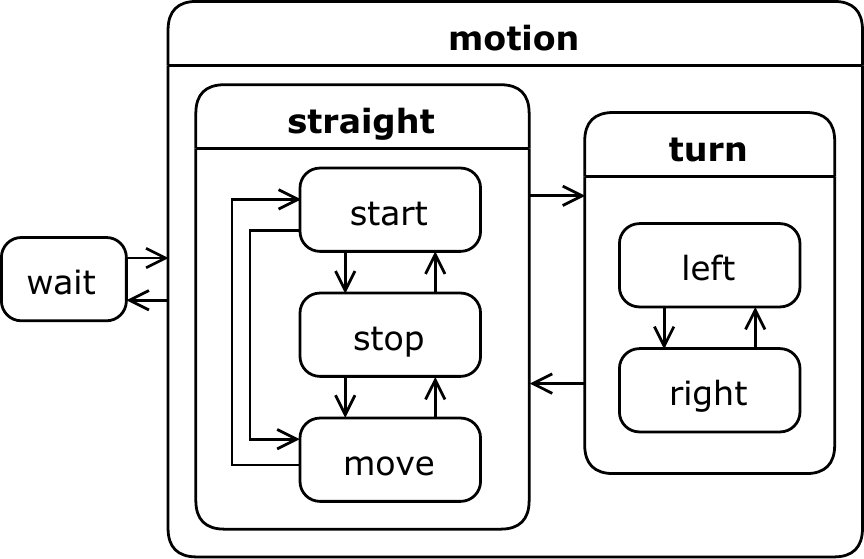}
		\vskip 0mm
		\caption{Cyclist motion state machine.}
		\label{fig:state_machine}
	\end{center}
	\vskip 0mm
\end{figure}

The overall goal of basic movement detection is to detect motion states of cyclists during which they show similar behavior. Using these detected states, we aim to achieve more precise forecasts of the cyclists' future trajectories compared to forecasts without additional information about their current motion state. We grouped every sample of our dataset $\mathit{D}$ into sets of basic movements: 
	\textit{move}: cyclist moves with nearly constant velocity,
	\textit{stop}: cyclist slows down to a halt,
	\textit{wait}: cyclist stands still until \textit{start},
	\textit{start}: cyclist begins to move after \textit{wait}, and
	\textit{left/right}: cyclist makes a left/right turn.
The basic movements \textit{start}, \textit{stop}, and \textit{move} can be active simultaneously with \textit{left/right}. However, to train a unique trajectory forecast model for every basic movement, we resolve this by grouping the movements into \textit{wait/motion}, indicating whether the cyclist is waiting or moving and \textit{turn/straight}, indicating whether the cyclist is turning or going straight. We added three superstates \textit{motion}, \textit{turn}, and \textit{straight}. The complete dataset $\mathit{D}$ is divided into corresponding subsets as follows: \mbox{$\mathit{motion}=\{x|x\in \mathit{D} \land x\notin \mathit{wait}\}$}, \mbox{$\mathit{turn}=\{x|x\in \mathit{motion} \land (x\in \mathit{left} \lor x\in \mathit{right})\}$}, and \mbox{$\mathit{straight}=\{x|x\in \mathit{motion} \land x\notin \mathit{turn}\}$}. The corresponding state machine is visualized in Fig.~\ref{fig:state_machine}. Using this state machine, we avoid basic movements with multiple active states at the same time, by adding samples where the cyclist is turning exclusively to the \textit{turn} set, regardless of the current movement state in \textit{straight}. Instead of detecting one basic movement for every time step and choosing an according forecast model, we aim to assign a probability estimate $p_{s;t}$ for every basic movement in $S=\{\textit{wait}, \textit{start}, \textit{stop}, \textit{move}, \textit{left}, \textit{right}\}$ and use the resulting probabilities $p_{s;t}$ with $s \in S$ to weight the specialized forecast models.

\subsubsection{Generation of Motion History Images}
\label{subsec:basic_movement_detection_mhigeneration}

In this section, we describe the generation of the MHI used for basic movement detection. To generate an MHI, the silhouette of a cyclist including bicycle is detected over the past $M$ time steps. The color intensity values of the silhouettes are decayed over time, meaning silhouettes from earlier time steps become darker then silhouettes from later time steps, and combined into one image, encoding the past movements of the cyclists. Fig. \ref{fig:pipeline}b depicts the consecutive steps of the MHI generation.


In the first step, a region of interest (ROI) enclosing the cyclist is generated. The cyclist is detected in the image at the current time $t$ using a pretrained Mask R-CNN with Inception ResNet \cite{tf_model_zoo}. The output of the network consists of object boxes and instance segmentation masks for different classes. To create a bounding box for the cyclist, the detected boxes for the classes \textit{person} and \textit{bike} are used. The detected boxes are associated to a cyclist using a simple nearest neighbor strategy. The \textit{person} box is combined with the \textit{bike} box to create a bounding box enclosing the cyclist including the bicycle. To account for movements of the cyclist from previous time steps, the size of the resulting bounding box is enlarged by a factor $f_b$ to create the ROI. The instance segmentation masks of the past time steps are saved as binarized two channel images $\text{I}_{seg;t}(u, v, c)$ to generate the MHI. The first channel of $\text{I}_{seg;t}(u, v, c)$ contains the instance segmentation of the class \textit{person}, the second contains \textit{bicycle}, respectively.

In the second step, the cyclist instance segmentations of the past $M$ steps are extracted from the segmentation images $\text{I}_{seg;t}(u, v, c)$ using the ROI created in the first step and they are resized to the width and height $W$ and $H$ of the MHI, resulting in a sequence of cyclist and bike masks within the ROI $I_{\text{MHI};t}(u, v, c)$.
The color intensity values of the detected masks are decayed using the decay value $\tau=\frac{M-i}{M}$, with $i$ being the $i$-th time step, and are stacked to an MHI using Algorithm \ref{algo_mhi}. To distinguish between movements of the cyclists and movements of the bicycle, we use a two channel MHI, with one channel containing the MHI of the \textit{person} and one channel containing the MHI of the \textit{bicycle} silhouette. The resulting MHI is depicted in Fig. \ref{fig:pipeline}b. This process is performed for both cameras resulting in $\text{MHI}_{1;t}(u, v, c)$ and $\text{MHI}_{2;t}(u, v, c)$.

\begin{algorithm}
	\caption{MHI Generation}
	\begin{algorithmic}[1]
		\State $I_{\text{MHI};t}(u,v,c) \gets$ cyclist/bike mask sequence at current time t, with pixel positions $u$, $v$, and image channels $c$
		\State $M \gets$ number of time steps encoded in ${MHI}_t(u,v,c)$
		\State $W \gets$ image width
		\State $H \gets$ image height
		\State $\Delta t:=\ $\SI{0.02}{\second} $\gets$ duration between time steps
		\State $C=2 \gets$ number of channels
		\State $MHI_t(u, v, c):=0$
		\For{$i=M-1\text{ to } 0$} \raggedleft\Comment{iterate over all images, \\start with oldest\space\space\space\space\space\space\space\space\space\space}
		\State $\tau(i)=\frac{M-i}{M}$\Comment{calculate decay value $\tau$} \raggedright
		\For{$c=0\text{ to } C-1$}\Comment{repeat for both channels}
		\For{$u=0\text{ to } W-1$}\Comment{go through all pixels}
		\For{$v=0\text{ to } H-1$}
		\If{$I_{s;t-(i\cdot\Delta t)}(u, v, c)==1$}	
		\State $MHI_t(u, v, c)=$ \Comment{update MHI}
		\State\,\,\,\,\,$\tau(i)\cdot I_{s;t-(i\cdot\Delta t)}(u, v, c)$
		\EndIf
		\EndFor
		\EndFor		
		\EndFor
		\EndFor
	\end{algorithmic}
	\label{algo_mhi}
\end{algorithm}

\subsubsection{Network Architecture}
\label{subsec:basic_movement_detection_networkarch}

In this section, the method to detect basic movements using MHI is described. Rather than predicting one vector containing probabilities for each basic movement, we use four different models to predict the probabilities for \textit{wait/motion} $\vec{p}_{wm}=\left[p_{wait}, p_{motion}\right]$, \textit{straight/turn} $\vec{p}_{st}=\left[p_{straight}, p_{turn}\right]$, \textit{left/right} $\vec{p}_{lr}=\left[p_{left}, p_{right}\right]$, and \textit{start/stop/move} $\vec{p}_{ssm}=\left[p_{start}, p_{stop}, p_{move}\right]$, to generate a model of the state machine in Fig. \ref{fig:state_machine}. By using separate models, we are able to train specialized classifiers for each part of the state machine.

The specialized classifiers are trained using MHIs from both cameras in combination with the past cyclist trajectory containing the last 50 positions (i.e., \SI{1}{\second}, see Sec. \ref{subsec:dataset}) as network input and the corresponding ground truth vector $\vec{l}_{wm;t}=\left[l_{wait}, l_{motion}\right]$, $\vec{l}_{ts;t}=\left[l_{turn}, l_{straight}\right]$, $\vec{l}_{lr;t}=\left[l_{left}, l_{right}\right]$, or $\vec{l}_{ssm;t}=\left[l_{start}, l_{stop}, l_{move}\right]$ containing the basic movement labels (zero or one) is used as network output. 

\begin{figure}
	\begin{center}
		\vskip 0mm
		\includegraphics[width = 0.9\columnwidth]{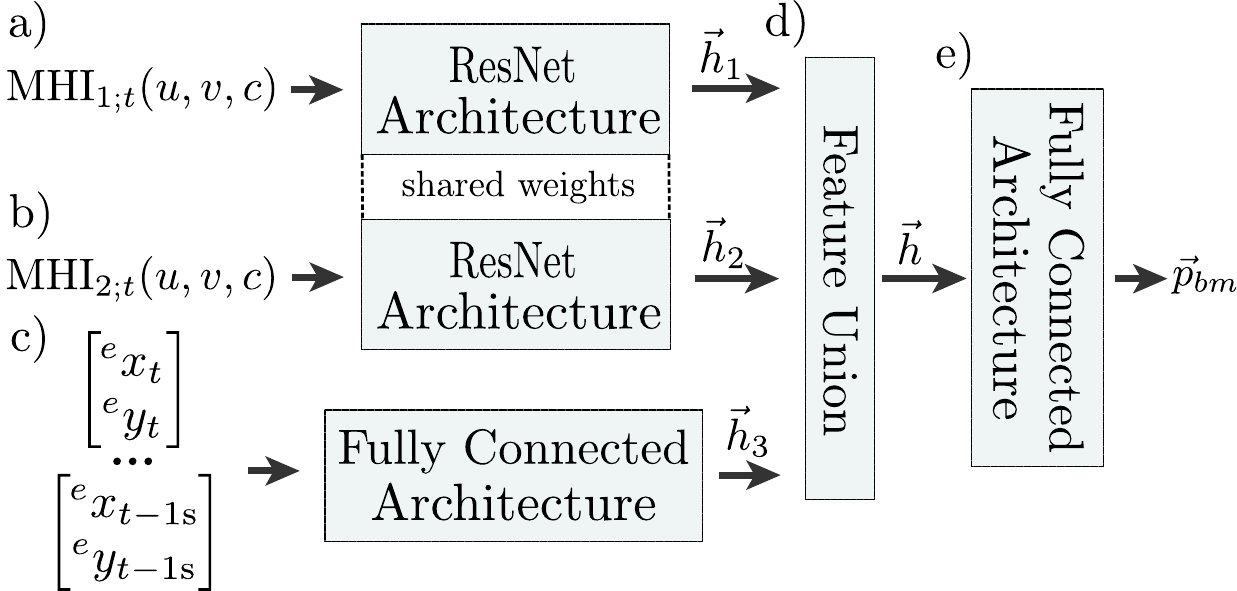}
		\vskip 0mm
		\caption{Basic movement detection network architecture.}
		\label{fig:bmd_arch}
	\end{center}
	\vskip -3mm
\end{figure}

Fig. \ref{fig:bmd_arch} describes the architecture of the classification networks. Both MHI are fed into CNN architectures (Fig. \ref{fig:bmd_arch}a and b). Since the ResNet architecture has proven to perform well in image classification tasks \cite{he_resnet}, we decided to use ResNet as CNN architecture. The past trajectory is fed into a fully connected architecture (Fig.~\ref{fig:bmd_arch}c). The outputs $\vec{h}_1$, $\vec{h}_2$, and $\vec{h}_3$ are concatenated (Fig.~\ref{fig:bmd_arch}d) and fed into another fully connected architecture (Fig.~\ref{fig:bmd_arch}e) followed by a fully connected layer with linear activation with its output size being the number of probabilities in the output vector $\vec{p}_{bm;t}$ and a softmax layer to generate the output probabilities, so that $||\vec{p}_{bm;t}||_1=1$. With $bm \in \{wm, st, lr, ssm\}$ describing the sub state machine with states $S_{wm}=\{\mathit{wait}, \mathit{motion}\}$, $S_{st}=\{\mathit{straight}, \mathit{turn}\}$, $S_{lr}=\{\mathit{left}, \mathit{right}\}$, and $S_{ssm}=\{\mathit{start}, \mathit{stop}, \mathit{move}\}$. To be able to use the predicted values as weights for specialized trajectory forecasts, we need to estimate probabilities, which we achieve by applying a cross-entropy loss function:
\begin{equation}
L_{ce,bm}(\vec{p}_{bm;t}, \vec{l}_{bm;t})=-\sum_{s\in S_{bm}}l_{s;t}\cdot\log(p_{s;t}),
\label{equ:cross-entropy-bmd}
\end{equation}with $\vec{l}_{bm;t}$ containing the ground truth values $l_{s;t}$ for every state $s$ in $S_{bm}$ and $\vec{p}_{bm;t}$ the predicted probabilities $p_{s;t}$, respectively. To improve the reliability of the predicted probabilities, we perform calibration using isotonic regression \cite{goodprobabilities}. Both calibrated and uncalibrated probabilities are evaluated.

\subsubsection{Evaluation Method}
\label{subsec:basic_movement_detection_evaluation}

To evaluate basic movement detection results, we look at the classification results of individual time steps. However, since the overall goal is to generate weights for specialized forecast models, we also have to evaluate the generated probabilities. 

To evaluate the classification results of individual time steps, we calculate the $\text{F}_{1}$-scores of the classifiers for every sub-state machine. Since we have multi class classification problems, we compute both, the micro averaged and macro averaged $\text{F}_{1}$-scores:
\setlength{\maxdepth}{0pt}
\begin{equation}
\text{micro-precision}=\frac{\sum_{c=1}^{N_c}tp_c}{\sum_{c=1}^{N_c}tp_c+\sum_{s=1}^{N_c}fp_s},
\label{equ:mic_prec}
\end{equation}
\begin{equation}
\text{micro-recall}=\frac{\sum_{c=1}^{N_c}tp_c}{\sum_{c=1}^{N_c}tp_s+\sum_{c=1}^{N_c}fn_c},
\label{equ:mic_recall}
\end{equation}
\begin{equation}
\text{macro-precision}=\frac{\sum_{c=1}^{N_c}\frac{tp_c}{tp_c+fp_c}}{N_c},
\label{equ:mac_prec}
\end{equation}
\begin{equation}
\text{macro-recall}=\frac{\sum_{c=1}^{N_c}\frac{tp_c}{tp_c+fn_c}}{N_c},
\label{equ:mac_recall}
\end{equation}

\begin{equation}
\text{F}_{1,\text{micro}}=\frac{\text{micro-precision}\cdot \text{micro-recall}}{\text{micro-precision}+\text{micro-recall}},
\label{equ:f1-score-micro}
\end{equation}
\begin{equation}
\text{F}_{1,\text{macro}}=\frac{\text{macro-precision}\cdot \text{macro-recall}}{\text{miacro-precision}+\text{macro-recall}},
\label{equ:f1-score-macro}
\end{equation}
where $N_c$ describes the number of classes, i.e., states in $bm$, $tp_c$ describes the number of true positives, $\textit{fn}_c$ and $fp_c$, the numbers of false negatives and false positives of the \mbox{$c$-th} class, respectively. Additionally, we visualize the results using confusion matrices to illustrate misclassification between classes.

To evaluate the quality of the generated probabilities, in addition to the log-likelihood (Eq. \ref{equ:cross-entropy-bmd}) we compute the Brier scores  for every state $s \in S$:

\begin{equation}
\text{BS}_{s}=\frac{1}{N_t}\sum_{i=1}^{N_t}(p_{s,i}-l_{s,i})^2,
\label{equ:brier-score}
\end{equation} with the predicted probability $p_{s,i}$ and label $l_{s,i}$ of state $s$ of the i-th sample of all $N_t$ test samples. In contrast to the log-likelihood, the Brier score is always between zero and one, with zero being the best possible outcome, making it easier to interpret. 
To visualize the reliability of the generated probabilities, we use Q-Q plots {\cite{chambers1983graphical}. Predicted probabilities are considered reliable, if a certain predicted probability $p$ leads to a correct classification in $p\cdot N$ times of all $N$ predictions, e.g., for predicted probabilities of 0.5, the predictions have to be correct in 50\% of all cases. The \mbox{Q-Q plots} are generated for each sub-state machine.

\subsection{Probabilistic Trajectory Forecast}

\label{subsec:probabilistic_trajectory_forecast}
This section covers the probabilistic forecast of cyclist trajectories. In Sec. \ref{subsec: probabilistic_trajectory_forecast}, we describe the model from \cite{ownZernetsch_iv2019}, on which the model in this work is based and which is also used as baseline model. In Sec. \ref{subsec: motion_specific_trajectory_forecast}, we extend this model to a framework for motion state specific forecasts and use the basic movement detection from Sec. \ref{subsec:basic_movement_detection} to properly model transitions between movement states.
Sec. \ref{subsec: evaluate_probabilistic_forecasts} describes various metrics for the evaluation of probabilistic forecasts.
\subsubsection{Probabilistic Forecast} \label{sec: prob_unimodal_forecasts}
\label{subsec: probabilistic_trajectory_forecast}
In \cite{ownZernetsch_iv2019}, a method which incorporates uncertainty estimates to deterministic trajectory forecasts by modeling uncertainties with normal distributions is proposed. Instead of forecasting only the most likely positions of a VRU, this method can be used to forecast how likely it is that a VRU will be within a certain region. These forecasts include the most likely positions at the means of normal distributions and uncertainty estimates encoded in confidence regions which in turn are described by covariance matrices. For reliable forecasts these confidence regions give us the probability that the VRU will reside within the region.
Since we use one parametric normal distribution per time horizon, we write the distribution as:
\begin{equation}
\mathcal{N}_{t+h}(\vec{y})  = \frac{exp(-\frac{1}{2} \overbrace{[\vec{y}-\vec{\mu}_{t+h}]^T \cdot \mathbb{S}^{-1}_{t+h} \cdot [\vec{y}-\vec{\mu}_{t+h}]}^{d_{t+h}^2})}{\sqrt{(2\pi)^{2} \det \mathbb{S}_{t+h}}},
\label{equ: Nego}
\end{equation}
with $\vec{\mu}_{t+h}=\begin{bmatrix}
\mu_{x;t+h} & \mu_{y;t+h}
\end{bmatrix}$ being the mean, $\mathbb{S}_{t+h}$ the covariance matrix for a forecasted time horizon $h \in H_{\text{forecast}}$ in longitudinal ($x$) and lateral ($y$) direction, and $\vec{y}$ being an arbitrary point in space $\mathbb{R}^2$.
They are used to calculate the squared Mahalanobis distance $d_{t+h}^2$ between the mean of the distribution to an arbitrary point $\vec{y}$.
The covariance matrices and the means are parameterized by a neural network. Therefore, they are functions of the input trajectory $^eT_{in;t}$ and the network parameters $W$. The covariance matrix can be constructed from the longitudinal and lateral standard deviations $\sigma_{x;t+h}$ and $\sigma_{y;t+h}$, as well as a correlation coefficient $\rho_{t+h}$:

\begin{equation}
\mathbb{S}_{t+h} = \begin{bmatrix}
\sigma^2_{x;t+h} & \rho_{t+h}\sigma_{x;t+h}\sigma_{y;t+h}\\
\rho_{t+h}\sigma_{x;t+h}\sigma_{y;t+h} & \sigma^2_{y;t+h}
\end{bmatrix}.
\label{equ: Sego}
\end{equation}

\begin{figure}
	\begin{center}
		\vskip 0mm
		\includegraphics[width = 0.9\columnwidth]{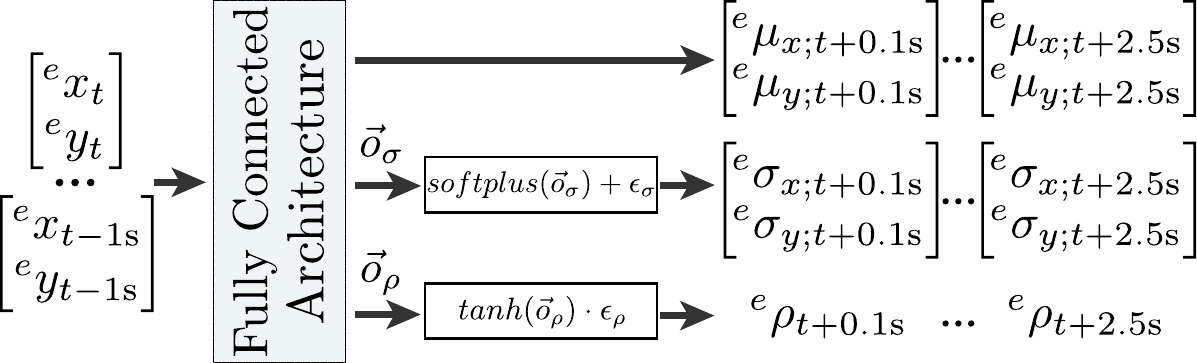}
		\vskip 0mm
		\caption{Specialized trajectory forecast network architecture.}
		\label{fig:traj_arch}
	\end{center}
	\vskip -5mm
\end{figure}

The covariance matrix is a symmetric and positive definite matrix. In the model implementation, the standard deviations $\sigma_{x;t+h}$ and $\sigma_{y;t+h}$ must be greater than zero and the correlation coefficient $\rho_{t+h}$ must be within $\left]-1, 1\right[$. 
Therefore, we have to transform the linear network outputs $\vec{o}_\sigma$ and $\vec{o}_\rho$} (Fig. \ref{fig:traj_arch}) which are used to form the covariance matrices $\mathbb{S}_{t+h}$ to match these constraints. For the correlation coefficient, we apply the hyperbolic tangens $tanh$ element-wise to transform the corresponding outputs $\vec{o}_\sigma$ into the desired interval. We also scale this interval to $\left]-\epsilon_{\rho},\epsilon_{\rho}\right[$ since a high correlation can lead to numerical problems.
In a preliminary investigation we found that the min/max values of the correlation coefficient, forecasted by the network are $\left[-0.75,0.75\right]$, therefore we see $\epsilon_{\rho}=0.9$ as a valid choice to model the distributions and suppress numerical instabilities. For the standard deviations we apply the Softplus-Function $\textit{softplus}(\vec{o}_\sigma) = \ln(1+e^{\vec{o}_\sigma})$ element-wise on the network outputs $\vec{o}_\sigma$ to obtain positive values. To prevent any $\sigma$ from becoming zero, we add a small $\epsilon_\sigma$ to the result of the Softplus-function. To learn the uncertainty estimates from the given data, we minimize the average negative log-likelihood over the forecasted time horizons, for every input and output trajectory pair, which can be seen as a cross entropy minimization. We decompose the negative log-likelihood of one forecasted time horizon into:

\begin{equation}
\begin{aligned}
\textit{NLL}_{t+h} &= \underbrace{\frac{1}{2} d_{t+h}^2(\vec{y}_{t+h},W,^eT_{in,t})}_{\textit{error}}\\ 
&+\underbrace{\frac{1}{2} \log \det\mathbb{S}_{t+h}(W,^eT_{in,t})}_\textit{residual}
\end{aligned}
\label{equ:NLL_uni}
\end{equation}
In this decomposition, the error term describes the distance between the forecasted distribution and a ground truth point $\vec{y}_{t+h}$ while the residual term contains a volumetric measure of uncertainty encoded in the covariance matrix. In an optimization scenario, the solver minimizes the error term first, since this can also be achieved with higher uncertainty estimates. The residual term keeps the uncertainty estimates as small as possible. Since the squared Mahalanobis distance $d_{t+h}^2$ within the error term is based on the matrix inverse and the residual term on the determinant of the covariance matrix $\mathbb{S}_{t+h}$, this loss is prone to numerical issues, especially during the gradient calculation. To resolve these issues, we invert the matrix and build the determinant using the Cholesky decomposition. As described in Sec. \ref{subsec:intention_detection}, we use trajectories in ego coordinates.
We optimize the network on \textit{NLL} averaged over all forcast horizons and all samples in the training set.
The forecasted distributions are given in ego coordinates. To obtain distributions in world coordinates, the geometric transformation described in Eq. \ref{equ:ego_transformation} must be reversed.

\subsubsection{Motion Specific Forecast}
\label{subsec: motion_specific_trajectory_forecast}
In \cite{ownZernetsch_iv2019}, it was discovered that a single normal distribution per forecast horizon is not sufficient to reliably model the forecast uncertainty.
It was concluded that the model tends to output underconfident forecasts. To reduce this effect, we provide a framework for motion specific forecasts in which every motion state is modeled by forecasted distributions of a specialized model:
\begin{equation}
\mathcal{D}_{t+h}(\vec{y}_{t+h}) = \sum_{s \in S} p_{s:t}\cdot \mathcal{F}_{s;t+h}(\vec{y}_{t+h}).
\label{equ:mixture_dist}
\end{equation} The joint distribution  $\mathcal{D}_{t+h}$ consists of the forecasted likelihood $\mathcal{F}_{s;t+h}$ of the individual basic movement $s$, weighted by a probability $p_{s;t}$ for the respective motion state.
For the basic movements \textit{start}, \textit{stop}, \textit{left}, \textit{right}, and \textit{move} the forecasted distributions are normal distributions $\mathcal{F}_{s;t+h}=\mathcal{N}_{s;t+h}$, parameterized by a neural network as described in Sec. \ref{sec: prob_unimodal_forecasts}, specialized for the respective motion state. Originally we handeled $\textit{wait}$ in the same way, but we found that the model cannot describe the variances for this motion state sufficiently.
Since $\textit{softplus}(x)$, which we apply to transform the outputs used for the standard deviation, decreases to zero exponentially for an $x$ smaller than one, we assume this to be a combination of a vanishing gradient problem and the fact that the future positions of \textit{wait} cannot be sufficiently modeled by a single Gaussian distribution.
For $\textit{wait}$ we fit a Gaussian mixture model \cite{murphy2012ml} for every time horizon $h$ of the forecasts $\mathcal{F}_{\textit{wait}:t+h} = \mathcal{GMM}_{\textit{wait}:t+h}$. We assume this to be valid since the input is always very similar for samples of the motion state $\textit{wait}$ and verified the validity with the methods of Sec. \ref{subsec: evaluate_probabilistic_forecasts}. The forecasts for the motion state $\textit{wait}$ are static and do not depend on the input trajectory $^eT_{in,t}$.

\subsubsection{Evaluation Methods}
\label{subsec: evaluate_probabilistic_forecasts}
As discussed in \cite{sharpness_nielsen}, we should focus on two characteristics of the forcasting model. One is the reliability and the second is the sharpness. While reliability describes whether the conditional variance of a forecast is correctly assessed in each situation, the sharpness measures the general narrowness of the forecasted distributions. Reliability is an essential property and sharpness can be used to decide how usable a model is in a real world application. For our application, we want to forecast distributions which are as sharp as possible, while retaining reliability, as they both serve as a basis for planning algorithms of autonomous vehicles.
In \cite{ownZernetsch_iv2019}, the authors evaluated the reliability of a forecasted distribution by plotting the confidence interval derived from the forecasts against the observed frequency of ground truth values within the confidence interval.
This method is not usable for arbitrary distributions, since the method used to calculate the confidence interval is only applicable for normal distributions. We propose the calculation of confidence sets and confidence levels instead of
intervals as a confidence measure for arbitrary distributions.
For a given distribution, we define a confidence set $\Omega(1-\alpha)$ as a set of points
with the confidence level of $1-\alpha$, with $\alpha\in\interval[open]{0.0}{1.0}$. 
To obtain the confidence level $1-\alpha(\vec{y})$ of a point $\vec{y}$, we search for points $\vec{z}\in\mathbb{R}^2$ which have a probability density level greater than or equal to the density $\mathcal{D}(\vec{y})$.

\begin{figure}
	\begin{center}
		\vskip 0mm
		\includegraphics[width = 0.7\columnwidth]{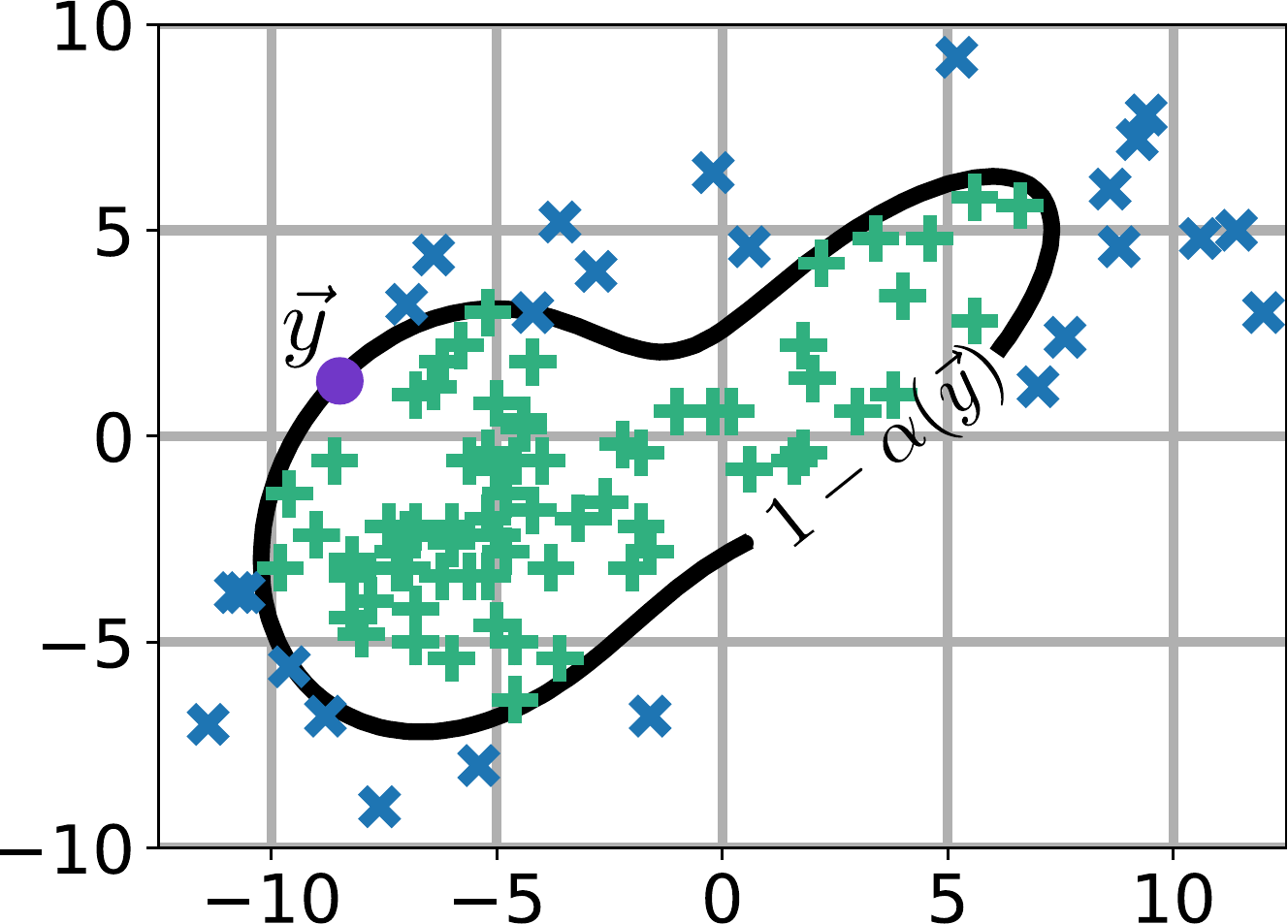}
		\vskip 0mm
		\caption{Confidence level $1-\alpha(\vec{y})$ (solid curve) of $\vec{y}$ (dot), with $N$ random samples $\vec{z}\in \mathbf{Z}$ with $\mathcal{D}(\vec{z})\geq\mathcal{D}(\vec{y})$ as + and $\mathcal{D}(\vec{z})<\mathcal{D}(\vec{y})$ as x.}
		\label{fig:distribution_eval}
	\end{center}
	\vskip 0mm
\end{figure}

Doing this for every point in $\mathbb{R}^2$ is not feasible in practice, so we draw $N$ random samples $\mathbf{Z}\sim\mathcal{D}$ from the distribution $\mathcal{D}$ to find these points. The confidence level estimate $1-\alpha(\vec{y})$ of the observation $\vec{y}$ is determined by:
\begin{equation}
1-\alpha(\vec{y}) = \frac{1}{N}\sum_{\vec{z}\in \mathbf{Z}}
\begin{cases}
1, & \text{ if }  \mathcal{D}(\vec{z}) \geq \mathcal{D}(\vec{y}) \\
0,              & \text{otherwise}
\end{cases}.
\label{equ:confidence_level}
\end{equation} Fig. \ref{fig:distribution_eval} illustrates the confidence level estimation for an exemplary distribution.
By knowing the confidence level of arbitrary points, we can build confidence sets $\Omega(1-\alpha)$ as a set of points with at least the confidence level $1-\alpha$:
\begin{equation}
\Omega(1-\alpha)=\big\{\vec{y}\vert\vec{y}\in\mathbb{R}^2\land\overbrace{1-\alpha(\vec{y})}^{\text{est. confidence level of }\vec{y}} \geq 1-\alpha\big\}.
\end{equation}
We can evaluate the reliability of the forecasting framework by calculating the estimated confidence level $1-\alpha(\vec{y})$ for every corresponding pair of forecasted distribution and ground truth point.
The observed relative frequency of occurrences $f_o(1-\alpha)$ should match $1-\alpha$ itself. We  evaluate the reliability visually by plotting the observed frequency over $1-\alpha$ which should ideally result in a line with a slope of one.
As comparative metrics we build $f_{o;t+h}(1-\alpha)$ from every corresponding pair of forecasted distribution $\mathcal{D}_{t+h}$ and ground truth position $\vec{y}_{t+h}$ in the validation set and search for the largest (Eq.~\ref{equ: murphy_rel_score}) and average (Eq.~\ref{equ: mean_rel_score}) distance from the ideal value $1-\alpha$ over all $\alpha \in A=\{0.01,0.02,...0.99\}$ and all time horizons $h \in H_{\text{forecast}}$:
\begin{equation}
\hat{\Gamma} = \max_{h}\max_{1-\alpha} |1-\alpha-f_{o;t+h}(1-\alpha)|
\label{equ: murphy_rel_score}
\end{equation}
\begin{equation}
\bar{\Gamma} = \frac{1}{|H_{\text{forecast}}| \cdot |A|}\sum_{h \in H_{\text{forecast}}}\sum_{\alpha \in A} |1-\alpha-f_{o;t+h}(1-\alpha)|
\label{equ: mean_rel_score}
\end{equation}
For these metrics the perfect score is 0, which is a completely reliable forecasting system, and the worst score is 1, which corresponds to an unreliable system.

For a certain confidence level $1-\alpha$ of a distribution we define the sharpness $\kappa(1-\alpha)$ as the volumetric measure of the confidence set $\Omega(1-\alpha)$. Since building the confidence set over $\mathbb{R}^2$ is analytically not feasible, a volumetric measure of the confidence sets and the confidence sets themselves are determined through quasi-Monte Carlo integration \cite{Morokoff95quasi-montecarlo}.

To provide a comparative metric for the sharpness of the forecasting system we build $\bar{K}(1-\alpha)$ as the average sharpness of a specific time horizon \mbox{$\bar{\kappa}_{t+h}(1-\alpha)$}, normalized to the forecast horizon:
\begin{equation}
\label{eq:ass}
\bar{K}(1-\alpha)=\frac{1}{|H_{\text{forecast}}|}\sum_{h\in H_{\text{forecast}}}\frac{\bar{\kappa}_{t+h}(1-\alpha)}{h}.
\end{equation}

To rate the positional accuracy, we use the Euclidean error between he most likely point of the forecasted distribution, i.e., the mode $\vec{\omega}_{t+h}$, and the ground truth $\vec{y}_{t+h}$ over every corresponding pair of forecasted mode and ground truth position and calculate the average Euclidean error $AEE_{t+h}$ for a specific time horizon over all pairs of forecasted modes and ground truth points. As a comparative metric, we use the ASAEE, introduced by Goldhammer \cite{goldhammer2019}, since it produces a single value of the forecast error normalized to the forecast horizon $H_{\text{forecast}}$:

\begin{equation}
\label{eq:asaee}
ASAEE=\frac{1}{|H_{\text{forecast}}|}\sum_{h\in H_{\text{forecast}}}\frac{AEE_{t+h}}{h}.
\end{equation}

\section{Experimental results}
\label{sec_ResultsOutline}


\subsection{Basic Movement Detection}
\label{subsec:results_basic_moevement_detection}
In this section, we present the results of the basic movement detection described in Sec. \ref{subsec:basic_movement_detection}. To train the models, the dataset described in Sec. \ref{subsec:dataset} was split into training, validation, and test sets using a 60-20-20 split. Since the samples were created from scenes and therefore, are not independent from each other, the split was performed by ensuring no samples from the same scene were assigned to different sets. Additionally, the samples were distributed to the sets in a way that each state $s$ has the same percentage in each set (as far as possible). To choose between calibrated and uncalibrated classifiers (Sec. \ref{subsec:basic_movement_detection_networkarch}), we created the \mbox{Q-Q}~plots for all classifiers using the calibrated and uncalibrated probabilities, chose the best classifiers using the validation set, and picked the classifiers with the best validation results. For \textit{wait/motion}, \textit{turn/straight}, and, \textit{left/right}, the calibrated classifiers showed the best validation results, for \textit{start/stop/move}, the uncalibrated classifiers showed better results. These classifiers were used to create the results in Sec. \ref{subsec:results_probabilistic_trajectory_forecast}. 

\begin{figure}
	\begin{center}
		\vskip 0mm
		\includegraphics[width = 0.99\columnwidth]{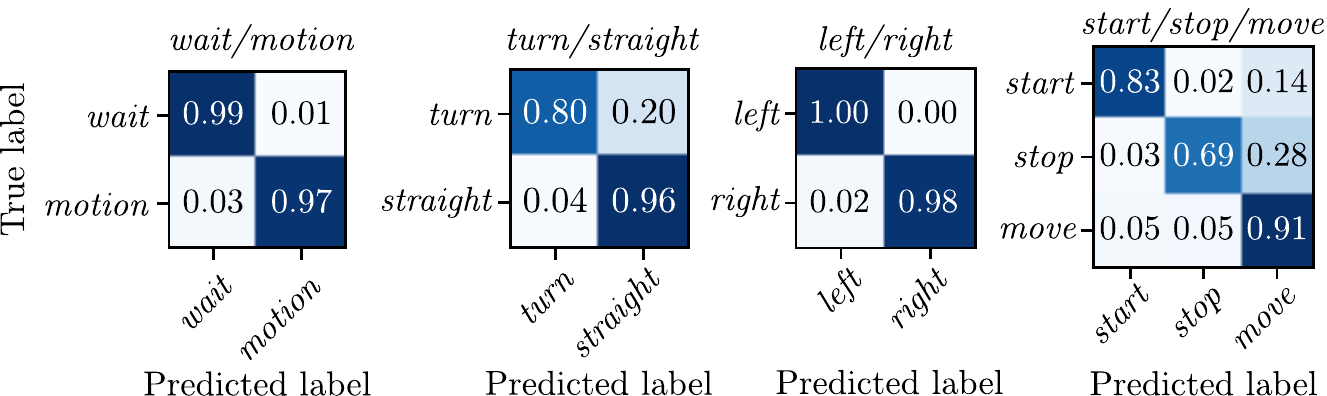}
		\vskip 0mm
		\caption{Confusion matrices of different sub-state machines.}
		\label{fig:bmd_confusion_matrices}
	\end{center}
	\vskip 0mm
\end{figure}

We first evaluate the classifiers' overall accuracy using the micro and macro $\text{F}_1$-scores (Tab.~\ref{tab:results_bmd}). Additionally, the confusion matrices of the best classifiers are depicted in Fig.~\ref{fig:bmd_confusion_matrices}. To compare the calibrated and uncalibrated classifiers, we created outputs for all classifiers using the test set. The \textit{wait/motion} and \textit{left/right} classifiers both perform very well. In both cases misclassifications mainly occur at the transitions between the two states. The confusion matrix of \textit{turn/straight} shows that turn samples are classified as \textit{straight} in 20\% of all cases. This often happens when the cyclist moves on a curve with a large radius, as well as at the beginning of a \textit{turn} movement where the past trajectory and the MHI show only few signs of a turning movement. Misclassifications of \textit{straight} samples mostly occur at the end of \textit{turn} movements, where past trajectory and MHI still show signs a of turn. By looking at the confusion matrix of \textit{start/stop/move}, we can see that \textit{start} and \textit{stop} are often predicted as \textit{move}. Especially the beginning of \textit{stop} is hard to classify, since \textit{stop} is only labeled if the cyclist slows down to a halt. Other breaking maneuvers are labeled as \textit{move}. If we compare the $\text{F}_1$-scores of the calibrated and uncalibrated classifier, we see only small differences.

\begin{figure}
	\begin{center}
		\vskip 0mm
		\includegraphics[width = 0.99\columnwidth]{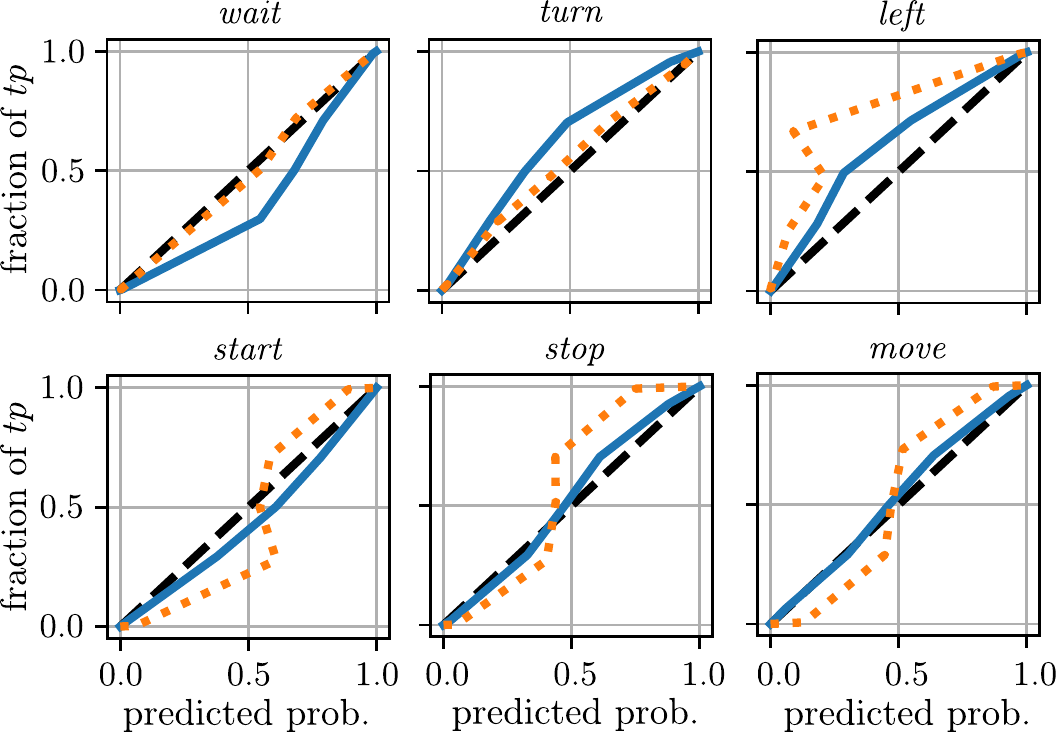}
		\vskip 0mm
		\caption{Q-Q plots of basic movement detection, with uncalibrated results (solid), calibrated results (dotted), and ideal results (dashed).}
		\label{fig:bmd_qq_plots}
	\end{center}
	\vskip 0mm
\end{figure}

Since the predicted probabilities are used to weight between specialized forecasting models, in addition to the overall accuracy of the predicted classes, we also have to evaluate the probabilities themselves. Therefore, we created the BS (Tab.~\ref{tab:results_bmd}) and \mbox{Q-Q plots} (Fig.~\ref{fig:bmd_qq_plots}) for the test set to rate the accuracy and reliability of the probabilistic predictions. Since \textit{wait/motion}, \textit{turn/straight}, and \textit{left/right} are two-class classifications with $p_1 = 1-p_2$, we only display the \mbox{Q-Q plots} for one of the classes, since the Q-Q plot of the second class results in the inverted plot of the first class. In the cases of \textit{wait/motion} and \textit{turn/straight} the calibration leads to slight improvements in Brier scores and reliability. Although the calibration of \textit{left/right} leads to an improvement on the validation set, the calibrated classifier shows worse results on the test set. This can be accounted to the fact that the \textit{left/right} classifier almost exclusively produces probabilities close to 1.0 or 0.0. Therefore, few probabilities can be used for the calibration. In the case of \textit{start/stop/move}, the calibration does not lead to an improvement for neither the validation nor the test set. However, the \mbox{Q-Q plots} of \textit{start}, \textit{stop}, and \textit{move} show that the results from the network are already well calibrated.

\begin{table}
	\caption{Classification results.}
	\begin{center}
	\vskip -2mm
	\scalebox{.76}{
	\begin{tabular}{|c|c|c|c|c|c|c|c|c|c|}
		\hline
		\multicolumn{10}{|c|}{Uncalibrated results}\\ \hline
		&\multicolumn{2}{c|}{\textit{wait/motion}}&\multicolumn{2}{c|}{\textit{turn/straight}}&\multicolumn{2}{c|}{\textit{left/right}}&\multicolumn{3}{c|}{\textit{start/stop/move}}\\ \hline
		$\text{F}_{1,\text{micro}}$&\multicolumn{2}{c|}{0.979}&\multicolumn{2}{c|}{0.927}&\multicolumn{2}{c|}{0.989}&\multicolumn{3}{c|}{0.853}\\ \hline
		$\text{F}_{1,\text{macro}}$&\multicolumn{2}{c|}{0.979}&\multicolumn{2}{c|}{0.883}&\multicolumn{2}{c|}{0.989}&\multicolumn{3}{c|}{0.820}\\ \hline
		&\textit{wait}&\textit{motion}&\textit{turn}&\textit{straight}&\textit{left}&\textit\underline{}{right}&\textit{start}&\textit{stop}&\textit{move}\\ \hline
		$\text{BS}_{s}$&0.016&0.016&0.054&0.054&0.008&0.008&0.057&0.058&0.097\\
		\hline
	\end{tabular}}
	\scalebox{.76}{
		\begin{tabular}{|c|c|c|c|c|c|c|c|c|c|}
			\hline \hline
			\multicolumn{10}{|c|}{Calibrated results}\\ \hline
			&\multicolumn{2}{c|}{\textit{wait/motion}}&\multicolumn{2}{c|}{\textit{turn/straight}}&\multicolumn{2}{c|}{\textit{left/right}}&\multicolumn{3}{c|}{\textit{start/stop/move}}\\ \hline
			$\text{F}_{1,\text{micro}}$&\multicolumn{2}{c|}{0.981}&\multicolumn{2}{c|}{0.933}&\multicolumn{2}{c|}{0.986}&\multicolumn{3}{c|}{0.853}\\ \hline
			$\text{F}_{1,\text{macro}}$&\multicolumn{2}{c|}{0.981}&\multicolumn{2}{c|}{0.886}&\multicolumn{2}{c|}{0.985}&\multicolumn{3}{c|}{0.820}\\ \hline
			&\textit{wait}&\textit{motion}&\textit{turn}&\textit{straight}&\textit{left}&\textit\underline{}{right}&\textit{start}&\textit{stop}&\textit{move}\\ \hline
			$\text{BS}_{s}$&0.015&0.015&0.050&0.050&0.012&0.012&0.073&0.077&0.128\\
			\hline
		\end{tabular}}
	\end{center}
	\label{tab:results_bmd}
	\vskip 0mm
\end{table}

\subsection{Probabilistic Trajectory Forecast}
\label{subsec:results_probabilistic_trajectory_forecast}

\begin{figure*}[!h]
	\begin{center}
		\vskip 0mm
		\includegraphics[width = 2.0\columnwidth]{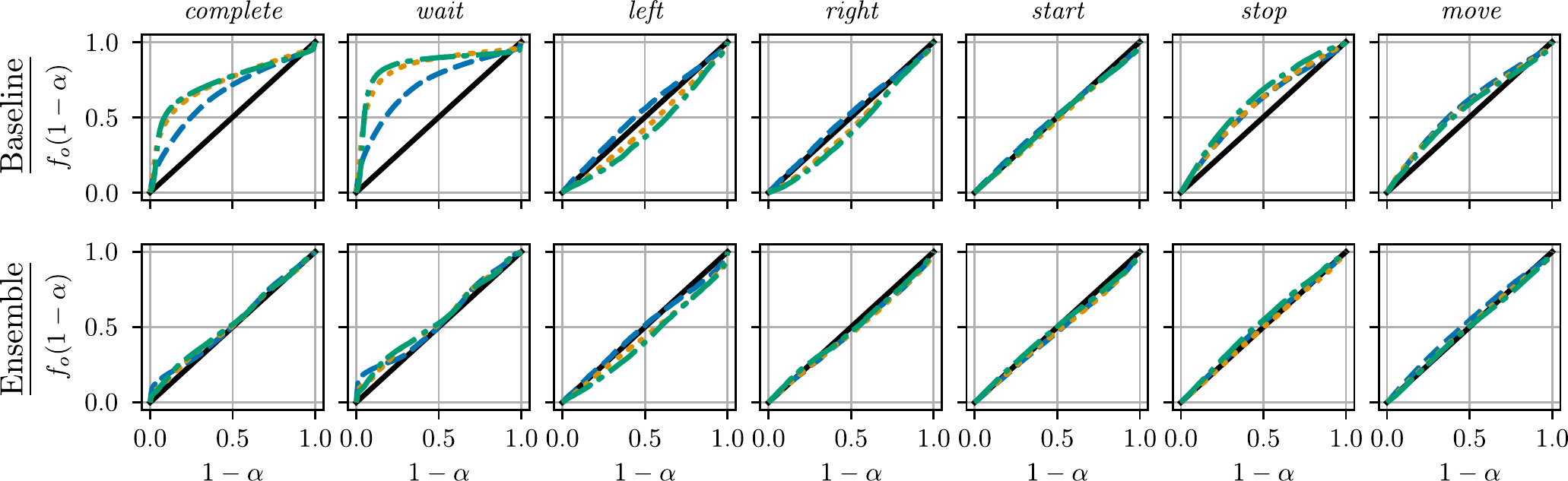}
		\vskip 2mm
		\caption{Reliability diagrams for the baseline (top row) and the ensemble (bottom row) for the forecasted time horizons \SI{0.5}{\second} (dashed), \SI{1.5}{\second} (dotted) and \SI{2.5}{\second} (dash-dotted).}
		\label{fig:reliability_diagram_ensemble}
	\end{center}
	\vskip 0mm
\end{figure*}

\begin{figure}
	\begin{center}
		\vskip 0mm
		\includegraphics[width = 1.0\columnwidth]{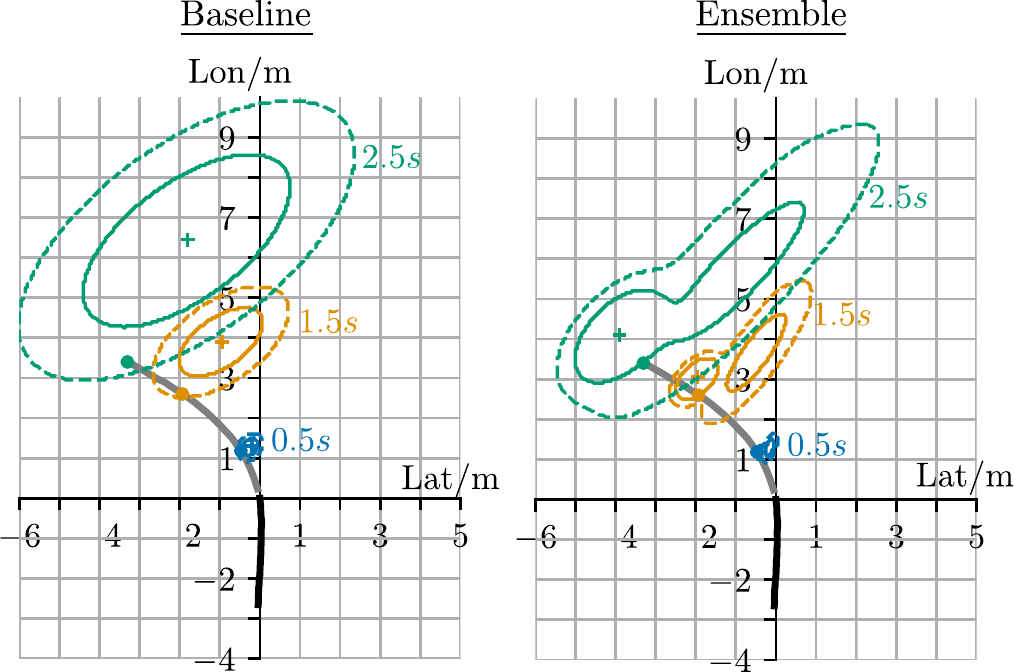}
		\vskip 0mm
		\caption{Example forecast from the baseline (left) and from the ensemble (right) approaches, with input trajectory (dark solid line), ground truth trajectory (gray solid line), 68\% levels (solid contours), 95\% levels (dashed contours), estimated modes (plus), and ground truth (dots on trajectory) at forecasted time horizons \SI{0.5}{\second}, \SI{1.5}{\second}, and \SI{2.5}{\second}.}
		\label{fig:ego_forecasts}
	\end{center}
	\vskip 0mm
\end{figure}

We trained one specific neural network for each motion state on data of \textit{start}, \textit{stop}, \textit{left}, \textit{right}, and \textit{move} each, as well as one general baseline neural network on data of a merged set from all motion states. For the motion state \textit{wait}, we fitted a Gaussian mixture model for every forecast time horizon. To train the models, the same split as in Sec. \ref{subsec:results_basic_moevement_detection} was used.

To select the best configuration for the models, a parameter sweep was performed. The models were trained for 500,000 training steps, where every 1,000-th step a validation was performed. The final networks were selected by choosing the configurations with the smallest validation score. We then trained the best network up to the step with the best validation score on a combined set of training and validation data.

\begin{table}
	\caption{Probabilistic trajectory forecast results.}
	\begin{center}
		\vskip -2mm
		\scalebox{.89}{
			\begin{tabular}{|c|c|c|c|c|c|c|c|}
				\hline
				\multicolumn{8}{|c|}{Baseline}\\ \hline

				&\textit{complete}&\textit{wait}&\textit{left}&\textit{right}&\textit{start}&\textit{stop}&\textit{move}\\ \hline
				$\hat{\Gamma}$&0.44&0.66&0.2&0.24&0.23&0.29&0.25\\
				$\overline{\Gamma}$&0.2&0.29&0.06&0.05&0.02&0.1&0.07\\
				\hline
				$\overline{K}(0.68)\text{ in }\si{m}^2/\si{s}$&$0.62$&$0.19$&$1.9$&$1.86$&$1.37$&$0.97$&$1.52$\\
				$\overline{K}(0.95)\text{ in }\si{m}^2/\si{s}$&$1.62$&$0.5$&$4.99$&$4.87$&$3.58$&$2.55$&$3.99$\\
				$\overline{K}(0.99)\text{ in }\si{m}^2/\si{s}$&$2.48$&$0.76$&$7.61$&$7.43$&$5.47$&$3.89$&$6.09$\\
				\hline
				$ASAEE\text{ in }\si{m}/\si{s}$&0.21&0.09&0.55&0.57&0.46&0.32&0.42\\

				\hline
		\end{tabular}}
		\scalebox{.89}{
			\begin{tabular}{|c|c|c|c|c|c|c|c|}
				\hline
				\multicolumn{8}{|c|}{Ensemble}\\ \hline
				
				&\textit{complete}&\textit{wait}&\textit{left}&\textit{right}&\textit{start}&\textit{stop}&\textit{move}\\ \hline
				$\hat{\Gamma}$&0.14&0.18&0.1&0.12&0.07&0.13&0.19\\
				$\overline{\Gamma}$&0.03&0.05&0.05&0.03&0.03&0.02&0.02\\
				\hline
				$\overline{K}(0.68)\text{ in }\si{m}^2/\si{s}$&$0.33$&$0.02$&$1.06$&$0.97$&$0.76$&$0.62$&$1.2$\\
				$\overline{K}(0.95)\text{ in }\si{m}^2/\si{s}$&$1.55$&$0.92$&$3.03$&$2.98$&$2.4$&$2.03$&$3.39$\\
				$\overline{K}(0.99)\text{ in }\si{m}^2/\si{s}$&$2.64$&$1.65$&$4.48$&$5.03$&$4.03$&$3.42$&$5.38$\\
				\hline
				$ASAEE\text{ in }\si{m}/\si{s}$&0.21&0.08&0.49&0.41&0.61&0.3&0.42\\

				\hline
		\end{tabular}}
	\end{center}
	\label{tab:results_traj}
	\vskip 0mm
\end{table}

Tab. \ref{tab:results_traj} shows the overall results of the probabilistic trajectory forecast. Fig. \ref{fig:reliability_diagram_ensemble} shows the reliability diagrams for the complete test set and for all movement types separately for the baseline (top row) and ensemble (bottom row) models. Compared to the baseline model, the ensemble achieved better overall scores in both largest $\hat{\Gamma}$ and average $\overline{\Gamma}$ distance to the ideal line (Fig. \ref{fig:reliability_diagram_ensemble}). $\hat{\Gamma}$ was reduced by 68\%, $\overline{\Gamma}$ by 85\%. While we see an improvement for all motion states, especially \textit{wait}, \textit{stop}, and \textit{move} show large improvements. The baseline produced underconfident forecasts, i.e., confidence regions are too wide compared to the ground truth, for these motion states. We attribute this to the use of a single Gaussian distribution per time step of the baseline, which is not sufficient to describe the real distribution. The sharpness evaluation for the \textit{complete} dataset shows smaller values for $\overline{K}(0.68)$ and $\overline{K}(0.95)$, and larger values for $\overline{K}(0.99)$ (see Tab. \ref{tab:results_traj}). However, except from the sharpness of \textit{wait}, the sharpnesses of all other motion states could be reduced by the ensemble, which is achieved through its multi-modality. The larger $\overline{K}(0.99)$ of \textit{wait} can be accounted to a small quantity of \textit{wait} samples \SI{2.5}{\second} before the beginning of \textit{start}, since these samples contain the movement of the \textit{start} motion in their ground truth trajectory. This behavior cannot be modeled by a single Gaussian distribution. The ASAEE of baseline and ensemble show similar results, while \textit{start} of the ensemble shows worse results compared to the baseline. Tests with ideal weights show an improvement of the ASAEE of the ensemble over the baseline for \textit{start} movements. Therefore, the larger ASAEE can be attributed to late \textit{start} detection. Fig. \ref{fig:ego_forecasts} shows an example forecast of the baseline model (left) and the ensemble (right) for a \textit{left} motion, showing a sharper forecast with better mode estimation for the ensemble.
\section{\large Conclusions and Future Work}
\label{sec:conclusion}

In this article, we presented a new approach to perform probabilistic cyclist intention detection. We compared our method to a baseline method and showed that our approach is able to achieve more reliable and sharper forecasts while retaining similar positional accuracies. When it comes to reliability we were able to reduce the average reliability error by 85\%. Sharpness was improved for all motion types except \textit{wait}.

In our future work, we will focus on improving basic movement detection, as it is crucial for the overall intention detection performance. In this work we assign a basic movement probability for every time step independently from previous probabilities. However, by incorporating knowledge about possible movement transitions in combination with detected states from previous steps, the accuracy and robustness of basic movement detection could be improved. We also plan to investigate ensemble methods to further improve basic movement detection.

To further improve the sharpness of our forecasts, we will investigate the use of different distribution functions for the specialized models, e.g., skew normal distributions or Gaussian mixtures. We also plan to compare our method to different approaches used to estimate probability distributions, such as Monte Carlo dropout \cite{mcdropout_gal} or quantile surfaces \cite{bieshaar2020quantile}.

Furthermore, we plan to validate our algorithm by incorporating probabilistic forecasts into an existing planning algorithm \cite{Eilbrecht.2017}.

\section{\large Acknowledgment}

This work results from the project DeCoInt$^2$, supported by the German Research Foundation (DFG) within the priority program SPP 1835: ``Kooperativ interagierende Automobile'', grant numbers DO 1186/1-2 and SI 674/11-2 and the project KI Data Tooling supported by the Federal Ministry for Economic Affairs and Energy (BMWi), grant numbers 19A20001L and 19A20001O. Additionally, the work is supported by ``Zentrum Digitalisierung Bayern''.



\bibliographystyle{IEEEtran}
\bibliography{sz}

\newpage

\begin{IEEEbiography}[{\includegraphics[width=1in,height=1.25in,clip,keepaspectratio]{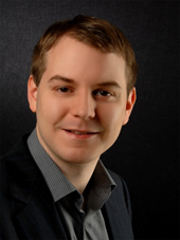}}]{Stefan Zernetsch}
received the B.Eng. and the M.Eng. degree in Electrical Engineering and Information Technology from the University of Applied Sciences Aschaffenburg, Germany, in 2012 and 2014, respectively. Currently, he is working on his PhD thesis in cooperation with the Faculty of Electrical Engineering and Computer Science of the University of Kassel, Germany. His research interests include cooperative sensor networks, data fusion, multiple view geometry, pattern recognition and behavior recognition of traffic participants.
\end{IEEEbiography}

\begin{IEEEbiography}[{\includegraphics[width=1in,height=1.25in,clip,keepaspectratio]{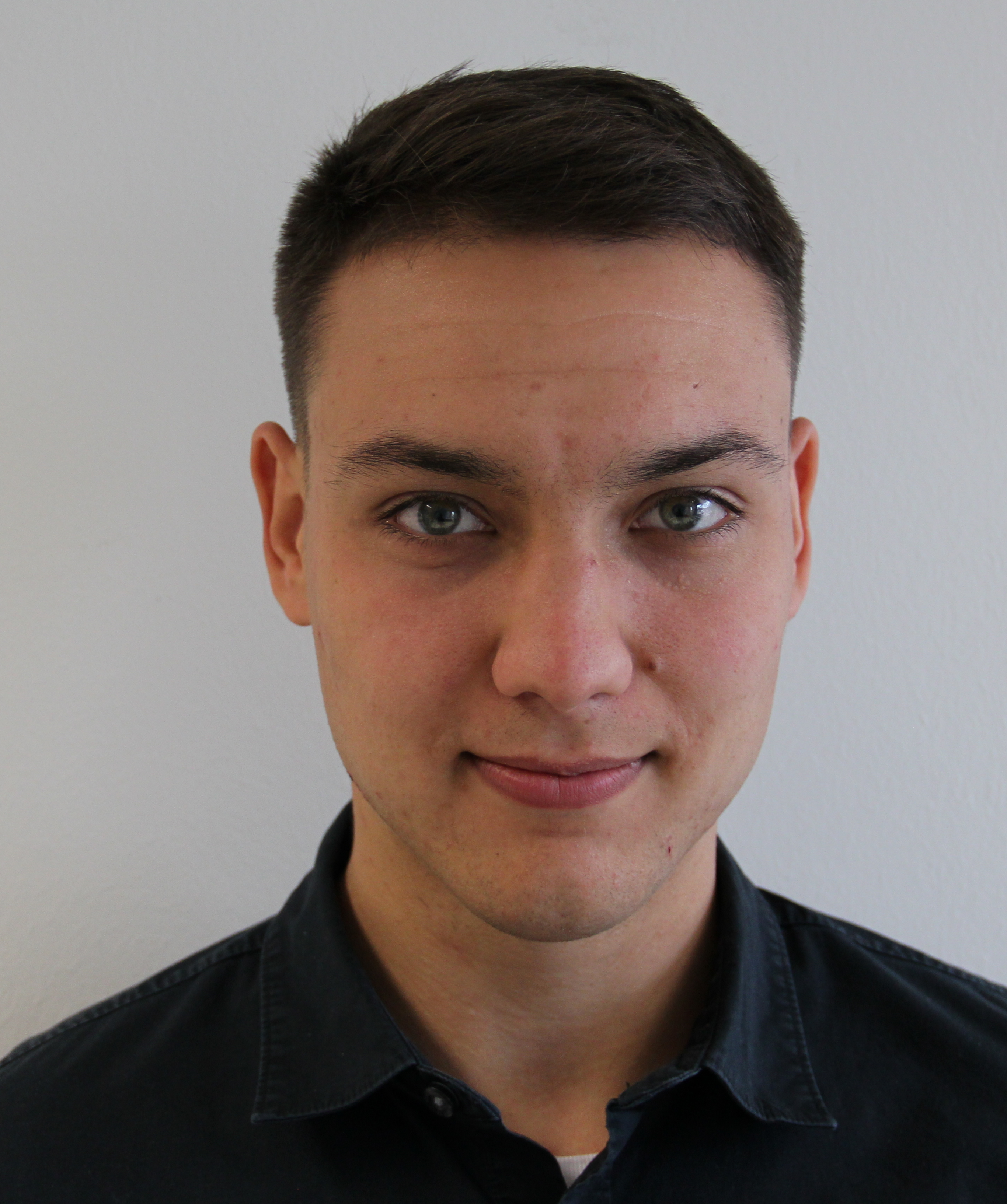}}]{Hannes Reichert} received the B.Eng. and the M.Eng. degree in Electrical Engineering and Information Technology from the University of Applied Seiences Aschaffenburg, Germany,	in 2018 and 2019, respectively. Currently, he is working as a research fellow at the University of Applied Seiences Aschaffenburg, Germany. His research interests include sensor data abstraction, data augmentation, computer vision, probabilistic modelling and and uncertainty analysis in machine learning models.
\end{IEEEbiography}
\vspace{-10mm}

\begin{IEEEbiography}[{\includegraphics[width=1in,height=1.25in,clip,keepaspectratio]{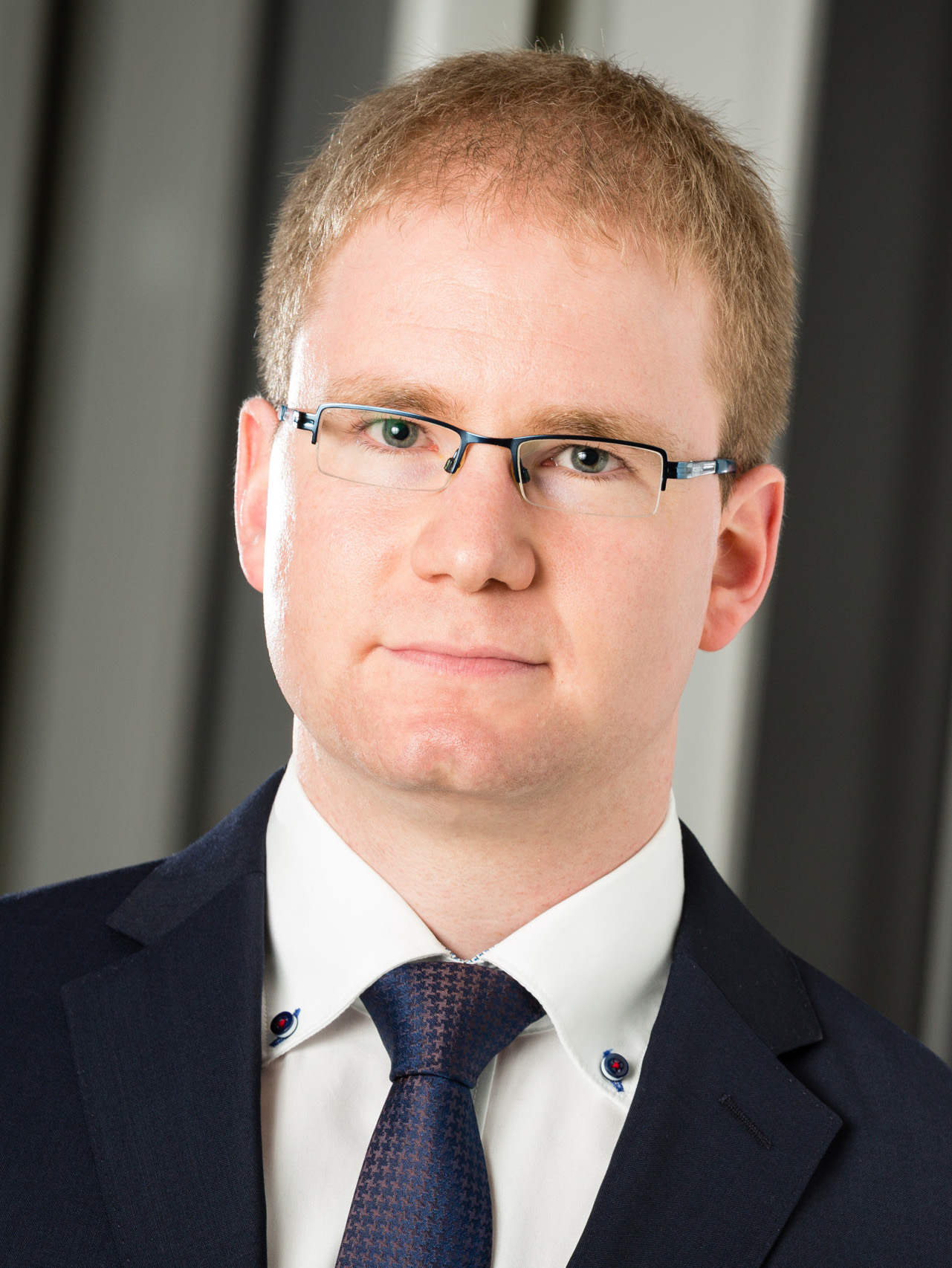}}]{Viktor Kress}
received the B.Eng. degree in Mechatronics and the M.Eng. degree in Electrical Engineering and Information Technology from the University of Applied Sciences Aschaffenburg, Germany, in 2015 and 2016, respectively. Currently, he is working on his PhD thesis in cooperation with the Faculty of Electrical Engineering and Computer Science of the University of Kassel, Germany. His research interests include sensor data fusion, pattern recognition, machine learning, behavior recognition and trajectory forecasting of traffic participants.
\end{IEEEbiography}
\vspace{-10mm}

\begin{IEEEbiography}[{\includegraphics[width=1in,height=1.25in,clip,keepaspectratio]{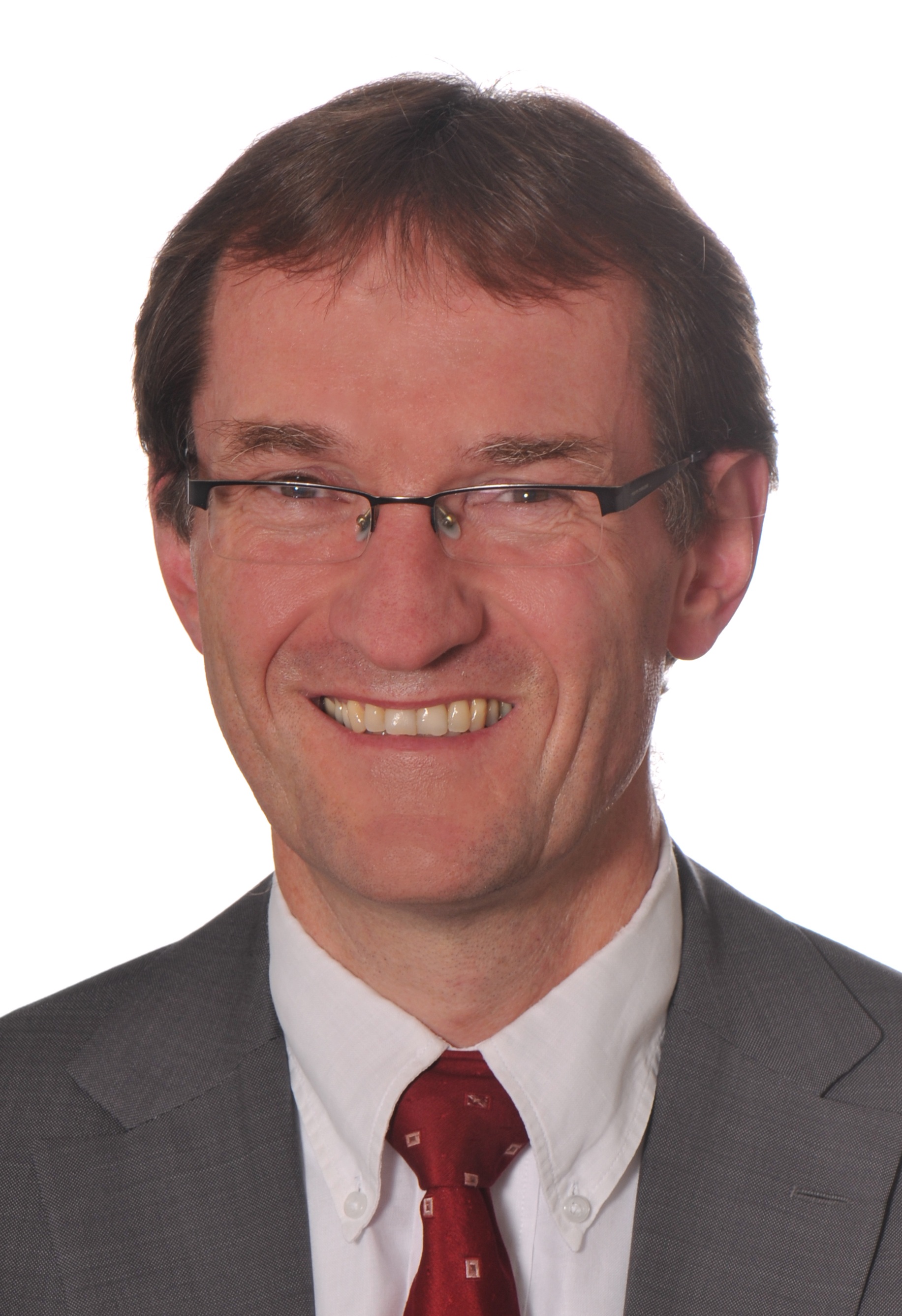}}]{Konrad Doll}
received the Diploma (Dipl.-Ing.) degree and the Dr.-Ing. degree in Electrical Engineering and Information Technology from the Technical University of Munich, Germany, in 1989 and 1994, respectively. In 1994 he joined the Semiconductor Products Sector of Motorola, Inc. (now Freescale Semiconductor, Inc.). In 1997 he was appointed to professor at the University of Applied Sciences Aschaffenburg in the field of computer science and digital systems design. His research interests include intelligent systems, their real-time implementations, and their applications in advanced driver assistance systems and automated driving. He received several thesis and best paper awards. Konrad Doll is member of the IEEE.
\end{IEEEbiography}
\vspace{-10mm}
\begin{IEEEbiography}[{\includegraphics[width=1in,height=1.25in,clip,keepaspectratio]{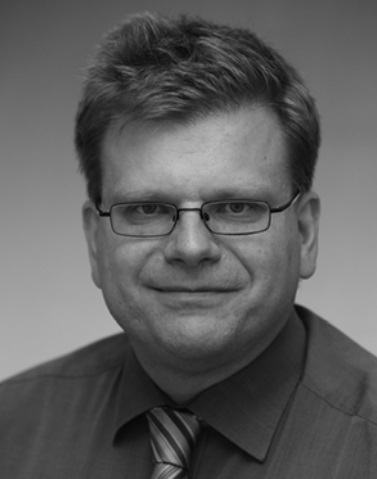}}]{Bernhard Sick}
received the diploma, the Ph.D. degree, and the "Habilitation" degree, all in computer science, from the University of Passau, Germany, in 1992, 1999, and 2004, respectively. Currently, he is full Professor for Intelligent Embedded Systems at the Faculty for Electrical Engineering and Computer Science of the University of Kassel, Germany. There, he is conducting research in the areas autonomic and organic computing and technical data analytics with applications in energy systems, automotive engineering, and others. He authored more than 180 peer-reviewed publications in these areas. Dr. Sick is associate editor of the IEEE TRANSACTIONS ON CYBERNETICS. He holds one patent and received several thesis, best paper, teaching, and inventor awards. He is a member of IEEE (Systems, Man, and Cybernetics Society, Computer Society, and Computational Intelligence Society) and GI (Gesellschaft fuer Informatik).
\end{IEEEbiography}
\vspace{-10mm}

\end{document}